\documentclass[letterpaper, 10 pt, conference]{ieeeconf}  %
\let\labelindent\relax

\IEEEoverridecommandlockouts                              %

\usepackage[pdftex]{graphicx}
\usepackage{svg}

\usepackage{caption}
\usepackage{subcaption}
\usepackage{amsmath}
\usepackage{amssymb}

\usepackage{algpseudocode}
\usepackage{algorithm}

\usepackage{multirow}
\usepackage{multicol}
\usepackage{enumitem}

\usepackage{gensymb}
\usepackage{bbm}

\usepackage{boldline}

\usepackage{tikz}
\usepackage{tikzpagenodes}
\usetikzlibrary{%
  arrows.meta
}

\definecolor{LinkColor}{HTML}{105FC1}
\usepackage{hyperref}
\hypersetup{
    colorlinks=true,   
    linkcolor=black,
    filecolor=black, 
    urlcolor=LinkColor,
    pdftitle={Online Self-Supervised Thermal Water Segmentation for Aerial Vehicles},
}

\title{\LARGE \bf
Online Self-Supervised Thermal Water Segmentation for Aerial Vehicles
}

\author{Connor Lee, Jonathan Gustafsson Frennert, Lu Gan, Matthew Anderson, and Soon-Jo Chung%
\thanks{*This work was supported by the Office of Naval Research. The authors are with the California Institute of Technology.
        {Email: clee@caltech.edu, ganlu@caltech.edu, matta@caltech.edu, sjchung@caltech.edu}.}
}

\begin{document}

\maketitle
\thispagestyle{empty}
\pagestyle{empty}

\begin{abstract}
We present a new method to adapt an RGB-trained water segmentation network to target-domain aerial thermal imagery using online self-supervision by leveraging texture and motion cues as supervisory signals. This new thermal capability enables current autonomous aerial robots operating in near-shore environments to perform tasks such as visual navigation, bathymetry, and flow tracking at night. Our method overcomes the problem of scarce and difficult-to-obtain near-shore thermal data that prevents the application of conventional supervised and unsupervised methods. In this work, we curate the first aerial thermal near-shore dataset, show that our approach outperforms fully-supervised segmentation models trained on limited target-domain thermal data, and demonstrate real-time capabilities onboard an Nvidia Jetson embedded computing platform. Code and datasets used in this work will be available at: 
\href{https://github.com/connorlee77/uav-thermal-water-segmentation}{https://github.com/connorlee77/uav-thermal-water-segmentation}. 
\end{abstract}

\section{INTRODUCTION}
\label{sec:intro}

Water segmentation is advantageous for uninhabited aerial vehicles (UAV) operating in near-shore environments. It can enable GPS-denied visual navigation~\cite{yang2017vision}, and assist tasks such as  bathymetry \cite{8726410}. However, current water segmentation algorithms operate on color (RGB) images and do not work well at night. Thermal cameras, on the other hand, can highlight details in conditions in which color cameras fail. In this paper, we look to develop a thermal water segmentation algorithm to bring autonomous nighttime capabilities to aerial robotics operating in near-shore settings like rivers, lakes, and coastlines.   

Compared to RGB water segmentation, which has been well studied in context of uninhabited surface vehicles (USV)~\cite{bb_iros_2019,yao2021shorelinenet,bovcon2018stereo}, thermal water segmentation has received little attention. As result, it lacks data, especially from aerial platforms, which prevents modern, state-of-the-art convolutional neural networks (CNN) from being easily applied. Moreover, three problems make it difficult to collect an aerial thermal near-shore data diverse enough for CNN training: Water bodies often coincide with no-fly zones; municipal-specific permits are required for non-recreational UAV usage; and distinct bodies of water are geographically dispersed, slowing diverse dataset collection for training and validation.

Aside from dataset limitations, thermal imagery is out-of-distribution relative to RGB imagery. As such, harnessing RGB data for thermal model training requires domain adaptation. Due to ongoing interest in self-driving cars, RGB-thermal domain adaptation has been well explored in urban settings~\cite{msuda, akkaya2021self, li2020segmenting, chen2022light, qian2020sparse, zhang2018synthetic}. However, such works still require training over target thermal data. Because we lack aerial, near-shore thermal data, we cannot effectively apply existing domain adaptation methods.

In this work, we propose a thermal water segmentation algorithm for UAVs that adapts to incoming thermal images during flight. \emph{Main contributions:} \textbf{1.} We present an online self-supervised approach that uses thermal water cues to adapt a RGB-pretrained water segmentation network to the near-shore thermal domain during flight. \textbf{2.} We demonstrate superior performance on aerial near-shore datasets compared to baselines. \textbf{3.} We present ablation studies of our self-supervision cues and test different RGB pretraining methods to assist online thermal adaptation. \textbf{4.} We release our algorithm as a Robot Operating System~\cite{ros} (ROS) package and demonstrate real-time online training and inference on a Nvidia Jetson AGX Orin. \textbf{5.} We release an annotated thermal water segmentation dataset, capturing aerial and ground near-shore settings, to bootstrap future work in this area.

\begin{figure*}
    \centering
    \includegraphics[width=\textwidth]{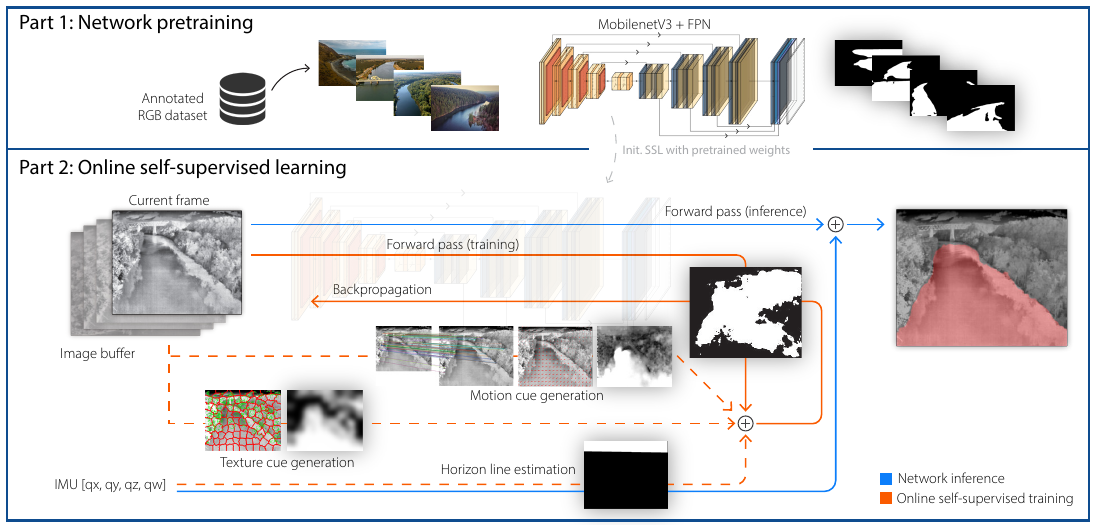}
    \caption{Schematic of our proposed online self-supervised training for thermal water segmentation. A few online training loops are performed prior to network inference on the current scene.}
    \label{fig:flowchart}
\end{figure*}

\section{Related Work}
\label{sec:relatedwork}

\textbf{RGB water segmentation: } 
RGB water segmentation methods typically leverage various combinations of water appearance, reflections, and location priors to segment water. To detect water for uninhabited ground vehicle (UGV) navigation, \cite{ rankin2006daytime} fuses color, texture, stereo range, and horizon line cues to systematically identify true water pixels. \cite{RankinWater} takes a similar approach but utilizes correlation between color and water reflections at predetermined distances to segment water. In river settings, \cite{meier2021river} exploits the shallow viewing angle of USVs to estimate and segment the river plane via water reflection symmetry. In contrast to these works, our aerial application precludes the use of water reflections as they are less prominent in the thermal domain and less accentuated at higher altitudes. We use texture cues and horizon line estimation in this work, but not color, as hue and saturation do not exist in thermal data.

Other water segmentation algorithms leverage geometric priors based on their target setting and vehicle. For USV navigation, \cite{achar2011self} trains an online self-supervised classifier to segment rivers, by assuming that shore regions above the horizon line are similar to shore regions below. Their assumption does not extend to our UAV setting, however, as the horizon line is not necessarily always in the frame due to aircraft pitch. \cite{kristan2015fast} proposes an obstacle segmentation algorithm for USVs in maritime environments, using the horizontal stacking of water, land/horizon, and sky as location priors for components in a Gaussian Mixture Model. However, these location priors are exclusive to USVs in maritime environments where shoreline generally isn’t visible on either sides and the camera is always near the water surface. ~\cite{bovcon2018stereo} extends this by incorporating inertial measurement unit-based (IMU) horizon line estimates into the location priors and enforcing stereo constraints for obstacle detection.

Recent deep learning-based approaches leverage small annotated RGB water datasets to train supervised CNN segmentation models~\cite{bb_iros_2019,yao2021shorelinenet, 9197194}. These methods are more robust and require less parameter tuning compared to earlier approaches such as \cite{ rankin2006daytime}, but require large, i.i.d datasets to properly train. As segmentation datasets are expensive and time-consuming to assemble, these works focus on maximizing generalization performance on existing, but limited, water segmentation datasets. 

\textbf{Thermal water segmentation: }
Little work has been done in this domain. RGB images have been used directly as input to train CNNs for thermal maritime obstacle segmentation~\cite{nirgudkar2021beyond} but were found to perform poorly compared to training on maritime thermal imagery~\cite{nirgudkar2023massmind}. To the best of our knowledge, this is the only public thermal water segmentation dataset that exist today and was released around the time of our work. However, this dataset targets USV maritime environments and does not yield good performance on our aerial near-shore data (Table~\ref{table:versus-trained}). As aerial thermal near-shore datasets do not yet exist, we opt for an online self-supervised approach in this work instead of these offline supervised methods. However, we do leverage such methods for network pretraining and further improve thermal segmentation performance via online self-supervised learning.

\textbf{RGB-thermal domain adaptation: } RGB-thermal (RGB-T) domain adaptation (DA) has been well studied in urban environments due to their applications to self-driving cars and have been used to harness large RGB datasets in conjunction with thermal data to train thermal networks. Generative methods like image translation have been used to automatically synthesize fake thermal imagery with labels from annotated RGB datasets for thermal model training~\cite{li2020segmenting, zhang2018synthetic}. However, they are known to hallucinate and introduce spatial structures that don’t appear in the target domain~\cite{isola2017image}. Unsupervised domain adaptation (UDA) methods like~\cite{msuda} and~\cite{gan2023unsupervised} seek to align RGB and thermal CNN features by training with shared weights and adversarial losses using annotated RGB data and unlabeled thermal data. Because public datasets of aerial thermal near-shore settings did not exist at the time of this work, we use collected aerial thermal near-shore data for validation and operate under the assumption of having no target domain data available. 

\textbf{Self-supervised learning (SSL): } In SSL, labels are automatically generated from data rather than from annotation. SSL is a broad topic and has been used in offline settings with applications including monocular road detection~\cite{5650300, dahlkamp2006self, 8486472}, terrain traversability~\cite{schmid2022self}, and general representation learning~\cite{chen2020simple, chen2021exploring}. It has also been used to adapt semantic segmentation networks to out-of-distribution data by enforcing augmentation consistency via a momentum network~\cite{araslanov2021self} which we leverage in our work. 

SSL can be applied online to adapt to new environments: For RGB river segmentation, \cite{achar2011self}, as previously discussed, uses visual cues with horizon line river priors to create training patches for online classifier training. \cite{zhan2019autonomous} creates training labels for a river segmentation network by assigning labels to unsupervised segmentation output based on the response of an onboard LiDAR sensor. \cite{daftry2018online} performs online SSL on lightweight CNNs using stereo information for ground plane segmentation and is most related to our work. In this work, we use online SSL to adapt an RGB-pretrained water segmentation network to the aerial thermal near-shore domain by generating labels from water texture and motion cues, and horizon line estimates.

\section{METHOD}
\label{sec:method}
We develop a thermal water segmentation method for UAVs that does not see thermal data prior to test time. Our method adapts an RGB-pretrained CNN segmentation model with online SSL to compensate for RGB-T covariate shift (Fig.~\ref{fig:flowchart}). Self-supervised labels are generated by exploiting texture and motion differences between land and water. To increase robustness, we utilize IMU-based horizon line estimation to remove false positive water pixels in the sky and use an alternative CNN-based sky segmentation when IMU data is corrupted or unavailable. We now outline our preprocessing procedure for 16-bit thermal images, RGB-based network pretraining method, self-supervised label generation process, and the online learning algorithm.

\subsection{Thermal Image Preprocessing}
\label{sec:method_preprocess}
Raw 16-bit thermal images are contrast stretched using the 1\textsuperscript{st} and 99\textsuperscript{th} percentile pixel values and followed by Contrast Limited Adaptive Histogram Equalization \cite{pizer1987adaptive}. In Sec.~\ref{sec:flow}, image pairs are stretched with the maximum of the 2\textsuperscript{nd} percentile and the minimum of the 98\textsuperscript{th}.

\subsection{Segmentation Network Pretraining}
\label{sec:method_pretrain}
We pretrain a segmentation network to speed up online training convergence. As UAVs are resource-constrained, we choose a compute-efficient Feature Pyramid Network (FPN) built on a MobileNetV3-small backbone with 2.3~million parameters~\cite{lin2017feature,howard2019searching}. The network takes in $1\times H\times W$ images and outputs $2 \times H \times W$ class probability maps.

We train the network using water-related RGB images from ADE20K~\cite{zhou2018semantic}, COCO-stuff~\cite{caesar2018coco}, and a river segmentation dataset \cite{8258373}. We supplement with Flickr images, found by querying keywords like \textit{aerial river} and \textit{drone ocean}, and annotated using an ADE20K-pretrained segmentation network from \cite{zhou2018semantic}, resulting in 14,240 training images. Annotations are converted into \textit{water} or \textit{non-water} classes.

As thermal images are single-channel, we transform 3-channel RGB images into 1-channel grayscale between $[0, 1]$ prior to training using one of these methods:
\begin{itemize}[leftmargin=*]
    \item[-] \textbf{Grayscale}: OpenCV's default RGB to grayscale conversion method \texttt{cv2.cvtColor(...)}. 
    \item[-] \textbf{Random mix}: Weighted channel-wise mean with randomly selected weights. Random inversion is applied to simulate thermal temperature inversion.
    \item[-] \textbf{Random mix (PCA)}: RGB channels are decorrelated via principle components analysis (PCA). The first 2 channels are randomly mixed, normalized, and randomly inverted.
    \item[-] \textbf{RGB2Thermal}: RGB images are translated to thermal using contrastive unpaired translation \cite{park2020cut} after training on the MassMIND thermal, MaSTr1325~\cite{bb_iros_2019}, and our RGB dataset.
\end{itemize}

\subsection{Self-Supervision from Texture Cues}
\begin{figure*}
    \centering
    \includegraphics[width=\textwidth]{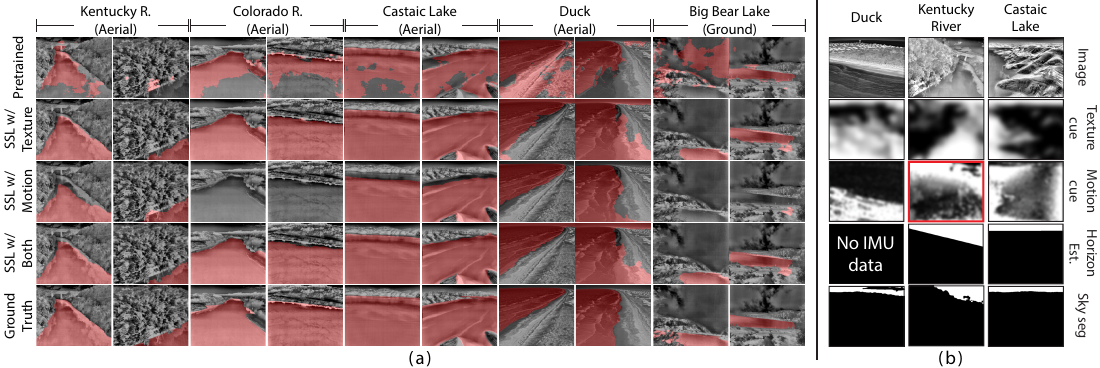}
    \caption{(a) Water segmentation results \textit{(red)} from a pretrained RGB network; our method with texture, and motion, and both cues; and ground truth. (b) Texture, motion, horizon, and sky segmentation cues used for online SSL. Motion cue failure case outlined in red.}
    \label{fig:segmentation_results}
\end{figure*}

Given image $I$, we create a soft water/non-water label for online SSL (Fig.~\ref{fig:segmentation_results}b) by observing that water tends to have less texture compared to surrounding land. We first perform unsupervised segmentation on $I$ via Simple Linear Iterative Clustering, creating superpixels similar in shape and size~\cite{achanta2010slic}. Each image pixel is assigned a class label based on the texture of the encompassing superpixel. We quantify texture using the Difference-of-Gaussian (DoG) keypoint detector~\cite{lowe1999object} and compute a probability map of non-water pixels 
\begin{equation}
    P_{\neg W}^T(i, j) = \left(G \circledast \frac{S_{kp}}{\alpha_T}\right)[i, j]
\end{equation}
by normalizing the keypoint count of each superpixel $S_{kp}$ with a parameter $\alpha_T$ and smoothing with a Gaussian kernel $G$. The probability of water pixels $P_{W}^T$ is the inverse ${1 - P_{\neg W}^T}$. Although the DoG detector filters out edge responses, we further mitigate jagged edge responses, such as along river banks, by pruning keypoints within 2 pixels of superpixel boundaries.

\subsection{Self-Supervision from Motion Cues}
\label{sec:flow}
In near-shore settings with fast flowing water like coastlines, water can appear choppy which breaks the assumption of the texture cue. However, this non-uniformity allows us to use water motion as another indicator of water pixels. We estimate water motion magnitude between successive image frames $I_{t-\Delta t}$ and $I_t$ using a two-step process:

First, we discount UAV-attributed motion by aligning successive frames using feature-based image registration. We match ORB features \cite{orb} detected in $I_{t-\Delta t}$ and $I_t$ and compute a homography matrix $H$ using Random Sample Consensus~\cite{ransac}. Because camera pose does not change significantly between successive frames, matches should consist mainly of shore features. $H$ is used to align the image coordinate frames via image warp $\mathcal{T}$ before cropping to the greatest common area. We then assume
\begin{equation}
    I_{t}^{\mathrm{crop}} \approx \mathcal{T}(I_{t - \Delta t}; H)^{\mathrm{crop}}
\end{equation}
effectively removing any UAV motion-induced transformations. We note that static shore regions must be in view in order to mitigate the risk of aligning based on water motion. 

Second, to create a probability map of water pixels $P_W^F(x, y)$, we quantify water motion using Farneback's algorithm~\cite{farneback2003two}, $\mathcal{F}$, to compute the dense optical flow field 
\begin{align}
    V\!=\! [V_x, V_y]^\top\!\! 
            =\! \mathcal{F}\!\left(I_{t}^{\mathrm{crop}},\!  \mathcal{T}(I_{t \!-\! \Delta t}; H)^{\mathrm{crop}}\right)
\end{align}
between the aligned image frame crops. We normalize the flow field magnitude by $\alpha_F$ to create a water probability map
\begin{equation}
    P_W^F(x, y) = \begin{cases}
        \frac{\|V(x, y)\|_2}{\alpha_F},  &\text{if } x \in [x_1, x_2], \ y \in [y_1, y_2] \\
        0,                       & \text{otherwise}
    \end{cases}
\end{equation}
and the non-water probabilities $P_{\neg W}^F$ arise as the inverse. We set $\alpha_F$ to be the 75\textsuperscript{th} percentile of the flow magnitudes.

\subsection{Discerning Sky and Water Pixels}
\label{method:sky-water-discern}
Sky and water may both have little texture, causing issues for texture cues. To fix this, we use horizon line estimation or sky segmentation to correct any resulting false positive water labels.

\subsubsection{Horizon Line Estimation with IMU Data}
\label{method:horizon}
Horizon (vanishing) line estimation allows us to unequivocally label all pixels above the horizon as non-water~\cite{bovcon2018stereo, achar2011self}. We estimate it by projecting distant points $\mathbf{x}_{\mathrm{u}}$ in the UAV coordinate frame that lay within the camera's field-of-view, to image coordinates $\mathbf{x}_{\mathrm{c}}$ via the relation
\begin{equation}
    \mathbf{x}_{\mathrm{c}} = \mathbf{P}_\mathrm{c} \mathbf{R}_{\mathrm{i}}^{\mathrm{c}} \ \mathbf{R}_{\mathrm{u}}^{\mathrm{i}} \mathbf{x}_{\mathrm{u}}
\end{equation}
where $\mathbf{R}_{\mathrm{u}}^{\mathrm{i}}$, $\mathbf{R}_{\mathrm{i}}^{\mathrm{c}}$ represent rotation matrices from UAV-to-IMU and IMU-to-camera, and $\mathbf{P}_\mathrm{c}$ is the camera projection matrix. We find the horizon by fitting a line to $\mathbf{x}_{\mathrm{c}}$ and take all pixels above to be \textit{non-water}. 

\subsubsection{Sky Segmentation without IMU Data}
\label{method:sky-segmentation-network}
In situations where IMU data is not accessible, we use a lightweight Fast-SCNN~\cite{poudel2019fast} segmentation network to quickly segment the sky. We remove the second bottleneck layer in the feature extractor to reduce computational burden. We train on MassMIND~\cite{nirgudkar2023massmind}, KAIST Pedestrian~\cite{hwang2015multispectral} with segmentation labels from~\cite{msuda}, SODA~\cite{li2020segmenting}, and FLIR aligned~\cite{ZHANG_2020_ICIP} data after reducing annotations to \textit{sky} and \textit{not-sky}.  

The FLIR aligned dataset does not have segmentation annotations. To create them, we segment FLIR RGB images using the same pretrained RGB network used to label Flickr images (Sec.~\ref{sec:method_pretrain}), creating sky masks. As the masks may be rough, and sky is \textit{usually} the coldest part in a thermal image, we refine each mask by binary searching the corresponding 14-bit thermal pixel values for a threshold that generates a new mask whose area falls within 10~\% of the RGB mask’s area. We visually inspect the results and retain 4,201 annotations out of 5,142 for sky segmentation training.

\begin{algorithm}
    \caption{Online Training and Inference}\label{alg:onlinetraining}
    {\fontsize{9}{10}\selectfont
    \begin{algorithmic}[1]
        \State \textbf{Input:} Network weights $\theta$, Image buffer length $L$
        \State \textbf{Output:} Segmentation masks $M_{0, 1, ... t}$
        \State \textbf{Initialize:} Networks $f_\theta$, $g_\theta$, Image buffer $Q$
        \\
        \While {camera on}
            \State Grab current image frame $I_t$
            \State Add $I_t$ to $Q$ and remove $I_{t-\Delta t L}$ if exists
            \\
            \If{train at time $t$}
                \State $\mathcal{D} \gets \textsc{CreateBatches}(Q)$
                \For{$n=1:N$}
                    \State Sample batch ($I_n, P_{W,n}^F, P_{W,n}^T)$ from $\mathcal{D}$
                    \State $P_n^g \gets g(I_n)$ 
                    \State $y_n \gets$ \textsc{MergeLabels}$(P_n^g , P_{W,n}^F, P_{W,n}^T)$ \Comment{Eq.~\ref{eq:merge1}-\ref{eq:merge2}}
                    \State $\theta_f \gets \theta_f + \gamma \nabla_{\theta_f}\mathcal{L}_{\mathrm{bce}}\left(f(I_n), y_n\right)$
                \EndFor
                \State $\theta_g \gets \lambda\cdot\theta_f + (1 - \lambda)\cdot\theta_g$ \Comment{Momentum update} 
            \EndIf
            \State $P_t^f \gets f(I_t)$ \Comment{Inference on current frame}
            \State Apply channel-wise argmax on $P_t^f$ to create $M_t$
            
            \State \textbf{yield} $M_t$
        \EndWhile
    \end{algorithmic}
    }
\end{algorithm}

\subsection{Online Training}
To perform online training (Algorithm \ref{alg:onlinetraining}), we initialize our pretrained segmentation network $f$ from Sec.~\ref{sec:method_pretrain}. We freeze the encoder and the first two decoder blocks to reduce trainable parameters. Like \cite{araslanov2021self}, we initialize a separate momentum network $g$ which is a copy of $f$. Network $g$ generates a soft self-label $P^g$ that is improved by ensembling with other cues and is updated with $f$'s weights at a rate of $\lambda=0.3$ after every training loop.

We adapt to the current scene by performing $N$ training iterations using images from a buffer that holds the past $L$ images seen, including the current image frame $I_t$. For each image in the buffer, we find the horizon line or sky segmentation depending on IMU availability, and generate our self-supervised labels. We merge the labels with $P^g$ using a per-class weighted average
\begin{align}
    \label{eq:merge1}
    y_{\mathrm{water}} &= w_1 P^g_W + w_2 P_W^F + w_3 P_W^T \\
    \label{eq:merge2}
    y_{\mathrm{non\hbox{-}water}} &= \xi_1 P^g_{\neg W} + \xi_2 P_{\neg W}^F + \xi_3 P_{\neg W}^T
\end{align}
and mark locations of sky pixels, or those above the horizon line, as definitively non-water. Network $f$ is trained on these soft labels using the binary cross entropy loss $\mathcal{L}_{\mathrm{bce}}$. 

During online training, images of size $512 \times 640$ are randomly cropped to $320 \times 320$ and subject to random horizontal flips. After $N$ training iterations, we perform inference on $I_t$ using $f$ to get $P^f$ and apply a channel-wise argmax to create segmentation mask $M_t$. We clean up the mask using morphological operations and keep the largest segmented contour as water. When IMU data is available, we also remove any water pixels still present above the horizon line. Lastly, we note that it is not necessary to perform online training prior to every inference call as image frames within a narrow time window are very similar.

\section{RESULTS}
\label{sec:result}

\subsection{Dataset}
\begin{table}[]
    \centering
    \caption{Thermal river, lake, and coastal datasets}
    \label{table:datasets}
    \resizebox{\columnwidth}{!}{
    \begin{tabular}{cccccc}
    \hlineB{3}
    \multirow{2}{*}{\textbf{Dataset}} & \multirow{2}{*}{\textbf{\shortstack[c]{Near-shore \\ Category}}} & \multirow{2}{*}{\textbf{\shortstack[c]{Capture \\ Method}}} & \multirow{2}{*}{\textbf{\# Images}} & \multirow{2}{*}{\textbf{\# Annot.}} & \multirow{2}{*}{\textbf{\# Seq.}} \\\\
    \hline
    \hline
    Kentucky River, KY  & River & UAV Flight & 7826 & 94 & 1 \\
    Colorado River, CA  & River & UAV Flight & 84,993 & 659 & 2 \\
    Duck, NC$^\dagger$    & Coast & UAV Hover & 4143 & 68 & 7 \\
    Castaic Lake, CA   & Lake & UAV Flight & 101,999 & 128 & 2 \\
    Big Bear Lake, CA   & Lake & Ground & 48,676 & 282 & 8 \\
    Arroyo Seco, CA     & Stream & Ground & 7 & 7 & --- \\
    \hlineB{3}
    \end{tabular}
    }
    \captionsetup{font=footnotesize}
    \caption*{$^\dagger$ Captured and stored in processed 8-bit data.}
\end{table}

\begin{figure}
    \centering
    \includegraphics[width=\columnwidth]{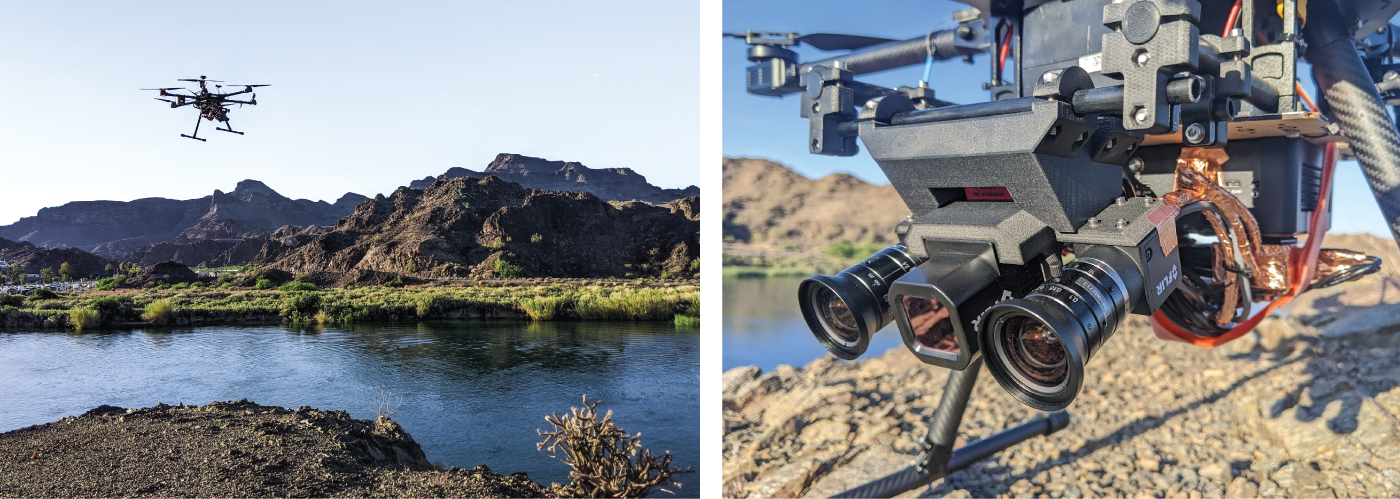}
    \caption{Our UAV operating over the Colorado River, CA, and the sensor stack mounted to the UAV showing the time-synchronized FLIR ADK thermal camera and VN100 IMU. Visible light cameras were not used as part of this work.}
    \label{fig:data-collection-overview}
\end{figure}

Our dataset consists of aerial and ground thermal sequences covering river, coastal, and lake scenery (Table \ref{table:datasets}) captured in 16-bit\footnote{The data from Duck, NC was captured in 8-bit format using a separate sensor stack and does not have IMU information available.} using a FLIR ADK long-wave thermal camera (Fig.~\ref{fig:segmentation_results},~\ref{fig:data-collection-overview}). These datasets are provided as ROS bag files, and also as individual frames with synchronized IMU and geolocation data for convenience. Frames were sampled for annotation at 2 second intervals, but at 12 second intervals for lengthy sequences from Castaic Lake. Some frames were skipped, at annotators' discretion, if indistinguishable to preceding frames. A single frame was used per Arroyo Seco sequence due to minimal change in each recording. 

Aerial sequences were used for experimental validation while ground sequences were used for training and ablations in non-target settings (see Table~\ref{table:near-shore-ablation} for list of sequences). Overall, the locations are very distinct and the datasets consist of a rich variety of sun positions, shore topography, water body size and shape, and surrounding flora. 
Aerial data from the Kentucky River (near Shakers Ferry Rd), KY; Colorado River (near Parker Dam), CA; Castaic Lake, CA; and the coastline at Duck, NC, were nominally recorded between 40 and 50~m above the water surface. Lower altitude imagery was also captured to enlarge the dataset for use in future work. Ground-level datasets from Big Bear Lake, CA and the Arroyo Seco (Pasadena), CA feature much shallower viewing angles of water and scenes with reduced visibility due to fog.

\subsection{Network Training Details}
The RGB-pretrained network from Sec.~\ref{sec:method_pretrain} and the sky segmentation network from Sec.~\ref{method:sky-segmentation-network} were trained as follows: Training images were resized to a longest dimension of 512, rescaled between 0.5 and 2.0, and randomly cropped to 320 $\times$ 320. Random horizontal flips, rotations (within 10\degree), and color jitter followed prior to single channel conversion if needed. We used stochastic gradient descent (SGD) with a momentum of 0.9, a learning rate of $1 \times 10^{-2}$, $L_2$ weight decay of $1\times 10^{-4}$, and a batch size of 32. 

Thermal-trained networks used as baselines in Sec.~\ref{sec:performance-eval} were trained using the same hyperparameters and data augmentations as above. However, 16-bit thermal images were first normalized using the thermal preprocessing technique described in Sec.~\ref{sec:method_preprocess}, but contrast stretched with random low and high values bounded by the 5\textsuperscript{th} and 95\textsuperscript{th} percentiles. 

\subsection{Online Training Setup}
\label{section:online-training-params}
For online training, we used the Adam optimizer with learning rate $1\times 10^{-3}$ and $L_2$ weight decay $1\times 10^{-4}$. Batch normalization was turned off. To increase speed, we scaled down images by 0.5 prior to label generation. By default, each round of online training ran for $N=8$ iterations with an 8 image buffer and a batch size of 4. Online training was performed every 120 frames and before each annotated frame, with every 4\textsuperscript{th} image added into the buffer.     

We set texture cue generation $\alpha_T = 10$ and set cue merging parameters $w = [1, 1, 1]$ and $\xi = [1, 0, 1]$. When a cue is not used, its corresponding weights are set to 0, e.g. $w = [1, 1, 0]$ and $\xi = [1, 0, 0]$ when only the motion cue is in use. We test networks initialized with weights via grayscale, random mixing, and PCA random mixing pretraining.     

\begin{table}
    \centering
    \caption{Performance evaluation of our online method in target aerial settings compared to fully-supervised networks trained with limited thermal data.}
    \label{table:versus-trained}
    \resizebox{\columnwidth}{!}{
    \begin{tabular}{lc|ccc}
    \hlineB{3}
    \multirow{2}{*}{\textbf{Method}} & \multirow{ 2}{*}{\textbf{\shortstack[c]{Training Set}}} & \multicolumn{3}{c}{\textbf{Aerial Test Setting mIoU}} \\ \cline{3-5}
                                     &                                                         & \textbf{River} & \textbf{Lake} & \textbf{Coast} \\ 
    \hline
    \hline
    MobilenetV3 + FPN & Arroyo Seco & 0.619 & 0.560 & 0.583 \\
    MobilenetV3 + FPN & Big Bear Lake & 0.687 & 0.526 & 0.638 \\
    MobilenetV3 + FPN & Big Bear Lake + Arroyo & 0.794 & 0.630 & 0.719 \\
    MobilenetV3 + FPN & Colorado River & --- & 0.745 & 0.436 \\
    MobilenetV3 + FPN & MassMIND~\cite{nirgudkar2023massmind}  & 0.454 & 0.310 & 0.445 \\
    \hline
    Online SSL (Grayscale) + TC & --- & \textbf{0.902} & 0.909 & 0.623 \\
    Online SSL (Grayscale) + MC & --- & 0.451 & 0.275 & 0.713 \\
    Online SSL (Grayscale) + All & --- & 0.885 & \textbf{0.911} & 0.668 \\
    
    \hline
    Online SSL (Rand. Mix) + TC & --- & 0.900 & 0.891 & 0.617 \\
    Online SSL (Rand. Mix) + MC & --- & 0.482 & 0.275 & 0.726 \\
    Online SSL (Rand. Mix) + All & --- & 0.884 & 0.904 & 0.659 \\
    
    \hline
    Online SSL (PCA) + TC & --- & 0.895 & 0.889 & 0.611 \\
    Online SSL (PCA) + MC & --- & 0.474 & 0.746 & \textbf{0.805} \\
    Online SSL (PCA) + All & --- & 0.878 & 0.909 & 0.654 \\
    
    \hlineB{3}
    \end{tabular}
    }
    \captionsetup{font=footnotesize}
    \caption*{TC -- Texture Cue \ \ \ \ \ \ MC -- Motion Cue}
\end{table}

\begin{table*}[]
    \fontsize{10}{12}\selectfont
    \centering
    \caption{Near-shore water segmentation ablation in different thermal sequences.}
    \label{table:near-shore-ablation}
    \resizebox{\textwidth}{!}{
    \begin{tabular}{c|c|cc|cc|ccc|ccc|ccc}
    \hlineB{4}
    \multirow{2}{*}{\textbf{Setting}} & \multirow{2}{*}{\textbf{Dataset Sequence}} & \multirow{2}{*}{\textbf{PT}} &  \multirow{2}{*}{\textbf{\shortstack[c]{PT + \\ Self-Train}}} & \multirow{2}{*}{\textbf{TC Only}} & \multirow{2}{*}{\textbf{MC Only}} & \multicolumn{3}{c|}{\textbf{w/o Sky Seg. nor Horizon Est.}} & \multicolumn{3}{c|}{\textbf{w/ Sky Segmentation}} & \multicolumn{3}{c}{\textbf{w/ Horizon Est.}} \\ \cline{7-15}
    
    & & & & & & \textbf{PT + TC} & \textbf{PT + MC} & \textbf{PT + All} & \textbf{PT + TC} & \textbf{PT + MC} & \textbf{PT + All} & \textbf{PT + TC} & \textbf{PT + MC} & \textbf{PT + All} \\
    \hline
    \hline
        \multirow{4}{*}{\shortstack[c]{Aerial \\ River}} 
        & Kentucky River 2-1 & 0.700 & 0.528 & 0.787 & 0.506 & 0.859 & 0.809 & 0.834 & 0.881 & 0.797 & 0.860 & \textbf{0.884} & 0.810 & 0.857 \\
        & Colorado River 1 & 0.500 & 0.453 & 0.796 & 0.476 & 0.894 & 0.295 & 0.881 & 0.897 & 0.295 & 0.884 & \textbf{0.898} & 0.295 & 0.886 \\
        & Colorado River 3 & 0.513 & 0.690 & 0.798 & 0.440 & 0.886 & 0.315 & 0.881 & 0.898 & 0.315 & 0.888 & \textbf{0.902} & 0.317 & 0.892 \\
    
        \cline{2-15}
    & \textbf{Avg. Seq. mIoU} & 0.571 & 0.557 & 0.794 & 0.474 & 0.880 & 0.473 & 0.865 & 0.892 & 0.469 & 0.877 & \textbf{0.895} & 0.474 & 0.878 \\
    \hlineB{4}
        
        \multirow{3}{*}{\shortstack[c]{Aerial \\ Lake}} 
        & Castaic Lake 2 & 0.324 & 0.241 & 0.830 & 0.521 & 0.901 & 0.701 & 0.911 & 0.804 & 0.227 & 0.826 & 0.886 & 0.703 & \textbf{0.918} \\
        & Castaic Lake 4 & 0.552 & 0.495 & 0.775 & 0.417 & 0.876 & 0.790 & 0.876 & 0.890 & 0.322 & 0.889 & 0.893 & 0.789 & \textbf{0.900} \\
        \cline{2-15}
    & \textbf{Avg. Seq. mIoU} & 0.438 & 0.368 & 0.802 & 0.469 & 0.889 & 0.746 & 0.893 & 0.847 & 0.275 & 0.857 & 0.889 & 0.746 & \textbf{0.909} \\
    \hlineB{4}

        \multirow{8}{*}{\shortstack[c]{Aerial \\ Coast}} 
        & Duck 4 & 0.799 & 0.915 & 0.366 & 0.541 & 0.448 & \textbf{0.933} & 0.469 & 0.499 & 0.693 & 0.499 & --- & ---  & ---  \\
        & Duck 5 & 0.347 & 0.854 & 0.674 & 0.870 & 0.792 & \textbf{0.931} & 0.859 & 0.375 & 0.500 & 0.456 & --- & ---  & ---  \\
        & Duck 6 & 0.519 & \textbf{0.842} & 0.500 & 0.683 & 0.506 & 0.832 & 0.562 & 0.460 & 0.601 & 0.517 & --- &   & ---  \\
        & Duck 10 & 0.743 & \textbf{0.782} & 0.489 & 0.551 & 0.457 & 0.740 & 0.461 & 0.493 & 0.552 & 0.498 &  & \textbf{No IMU} &  \\
        & Duck 13 & 0.300 & 0.260 & 0.546 & 0.164 & \textbf{0.579} & 0.429 & 0.573 & 0.578 & 0.429 & 0.572 & --- &  & --- \\
        & Duck 14 & 0.532 & 0.454 & 0.770 & 0.684 & 0.758 & \textbf{0.963} & 0.817 & 0.758 & \textbf{0.963} & 0.817 & --- & --- & --- \\
        & Duck 15 & 0.838 & 0.819 & 0.673 & 0.779 & 0.738 & 0.809 & 0.838 & 0.738 & 0.803 & \textbf{0.839} & --- & --- & --- \\
    
        \cline{2-15}
    & \textbf{Avg. Seq. mIoU} & 0.583 & 0.704 & 0.574 & 0.610 & 0.611 & \textbf{0.805} & 0.654 & 0.557 & 0.649 & 0.600 & --- & --- & --- \\
    \hlineB{4}
        \multirow{9}{*}{\shortstack[c]{Ground \\ Lake}} 
        
        & Big Bear Lake 23 & 0.430 & 0.423 & 0.251 & 0.352 & 0.278 & 0.406 & 0.295 & 0.429 & 0.433 & 0.436 & \textbf{0.785} & 0.759 & 0.783 \\
        & Big Bear Lake 27 & 0.638 & 0.687 & 0.355 & 0.518 & 0.408 & 0.399 & 0.431 & 0.653 & 0.485 & 0.703 & 0.713 & 0.378 & \textbf{0.749} \\
        & Big Bear Lake 30 & 0.471 & 0.424 & 0.317 & 0.472 & 0.345 & 0.374 & 0.353 & 0.506 & 0.382 & 0.518 & 0.611 & 0.413 & \textbf{0.613} \\
        & Big Bear Lake 34 & 0.643 & 0.702 & 0.436 & 0.391 & 0.504 & 0.696 & 0.511 & \textbf{0.857} & 0.705 & 0.834 & \textbf{0.857} & 0.524 & 0.842 \\
        & Big Bear Lake 37 & 0.479 & 0.495 & 0.313 & 0.420 & 0.307 & 0.475 & 0.308 & 0.597 & 0.457 & 0.584 & \textbf{0.743} & 0.465 & 0.710 \\
        & Big Bear Lake 40 & 0.741 & 0.820 & 0.532 & 0.368 & 0.652 & 0.430 & 0.662 & \textbf{0.854} & 0.456 & 0.851 & 0.818 & 0.457 & 0.819 \\
        & Big Bear Lake 44 & 0.756 & \textbf{0.846} & 0.376 & 0.400 & 0.289 & 0.772 & 0.375 & 0.830 & 0.828 & 0.802 & 0.824 & 0.843 & 0.832 \\
        & Big Bear Lake 50 & 0.642 & 0.630 & 0.280 & 0.356 & 0.318 & \textbf{0.719} & 0.340 & 0.555 & 0.683 & 0.569 & 0.609 & 0.697 & 0.661 \\
    
        \cline{2-15}
    & \textbf{Avg. Seq. mIoU} & 0.600 & 0.628 & 0.357 & 0.410 & 0.388 & 0.534 & 0.409 & 0.660 & 0.554 & 0.662 & 0.745 & 0.567 & \textbf{0.751} \\
    \hlineB{4}
    \end{tabular}
    }
    \captionsetup{font=footnotesize}
    \caption*{PT -- Base Pretrained Network \ \ \ \ \ \ TC -- Texture Cue \ \ \ \ \ \ MC -- Motion Cue}
\end{table*}

\subsection{Performance Evaluation}
\label{sec:performance-eval}
We demonstrate our method's robustness and superiority in a no-data regime over standard supervised segmentation with limited data. We compare our online SSL method against five thermal segmentation networks by testing in the primary aerial near-shore settings of this work: river, lake, and coast. These baselines were trained on the recently released MassMIND thermal USV segmentation dataset, the thermal ground-based data from Table~\ref{table:datasets}, and the aerial Colorado River dataset (Table \ref{table:versus-trained}). 

The baseline performances confirm our suspicions of poor generalization capabilities and overfitting due to limited dataset size, diversity, and covariate shift from surface/ground to aerial (Table~\ref{table:versus-trained}). Notably, neither networks trained on ground-level lake data (Big Bear Lake + Arroyo) nor networks trained on aerial river data (Colorado River) perform well when moving to aerial lake, and the lackluster performance of the MassMIND-trained network in these settings further motivates the collection and curation of aerial thermal datasets for nighttime UAV applications.     

In contrast, we report strong evidence favoring our texture- and motion-based online SSL over the fully-supervised networks: All three online variants using texture-based adaptation attain roughly 0.9 mIoU, outperforming the best thermal-trained networks in the aerial river (0.794 mIoU) and lake (0.745 mIoU) domains (Table~\ref{table:versus-trained}). Motion-based online adaptation performs best in the aerial coastal setting where wave motion and currents are highly visible. Here, the PCA-initialized variant outperforms the best thermal supervised network by a 0.08 margin, while the other two variants match performance. None of the motion-based variants perform well in rivers and lakes likely due to calmer waters. Likewise, texture-based cues do not perform well in coastal scenes due to confusion with highly-textured, fast-moving waves. Lastly, we see no significant advantages in leveraging both cues at the same time: river and lake settings see minor improvements while coastal settings see a performance drop. We finally note that these observations could be used to select suitable weights for cue merging (Eq.~\ref{eq:merge1}-\ref{eq:merge2}) during mission planning for operations in known near-shore environments. 

\subsection{Ablation Study}
\subsubsection{Influence of Online SSL} 

Using the overall best online model (PCA) from Sec.~\ref{sec:performance-eval}, we analyze the role of the texture and motion cues in the aerial near-shore settings (Table \ref{table:near-shore-ablation}). Overall, we find that cues do not provide adequate segmentation when used alone (Table \ref{table:near-shore-ablation}, TC/MC Only) and should be used to adapt a pretrained network as intended. They are necessary for online training robustness as self-training alone (PT+Self-Train) performs inconsistently. Our SSL cues, when used in appropriate online settings, i.e. texture-based with river/lake and motion-based with coastal, generally see mIoUs increase. Segmentation results with different SSL cues are displayed in Fig.~\ref{fig:segmentation_results}a.

To find the limits of our method, we perform further ablations on the ground-level Big Bear Lake sequences and find lower overall performance compared to aerial scenes (Table \ref{table:near-shore-ablation}). We attribute this to three things: First, a ground-level viewing angle from land causes water bodies to appear smaller, making the effect of noisy labels more pronounced. Second, water reflections tend to be more intense at shallower angles which texturizes water even when still. Lastly, dense fog and thermal sensor noise in some of sequences obscure the scene, making sky, background land, and water appear very uniform. Despite this, texture-based adaptation (PT+TC) still outperforms motion-based (PT+MC) by 0.11 mIoU with sky segmentation and 0.18 mIoU with IMU-based horizon estimation, reaffirming its use in calm water settings.

\subsubsection{Horizon Estimation and Sky Segmentation}
When IMU is available, horizon estimation can be used to boost segmentation performance (Table \ref{table:near-shore-ablation}). Moreover, as it does not rely on vision, it can mitigate the impact of fog and cloud obfuscations, as evident in the Big Bear Lake sequences where using the horizon yields over 0.35 gain in texture-based (PT+TC) mIoU versus having no knowledge of the sky or horizon. Sky segmentation via Fast-SCNN is less robust but still works well in the river settings and Castaic Lake 4. However, it is prone to mistaking far-field water as sky in Castaic Lake 2 and coastal scenes at Duck, leading to mIoU drop. It shows marked improvement over no sky segmentation in the Big Bear Lake scenes, demonstrating some robustness to fog, but still leaves room for improvement.  

\begin{table}
    \centering
    \caption{Pretraining method ablation using all thermal sequences.}
    \label{table:pretrain-ablation}
    \resizebox{\columnwidth}{!}{
    \begin{tabular}{c|cccc|c}
    \hlineB{3}
    \textbf{Pretraining Method} & \textbf{River} & \textbf{Lake} & \textbf{Coast} & \textbf{Ground} & \textbf{Avg. mIoU} \\
    \hline
    \hline
        Grayscale & 0.365 & 0.443 & 0.482 & 0.422 & 0.428 \\
        Rand. Mixing & 0.419 & 0.390 & 0.447 & 0.503 & 0.440 \\
        Rand. Mixing (PCA) & \textbf{0.563} & \textbf{0.441} & \textbf{0.642} & \textbf{0.574} & \textbf{0.555 }\\
        RGB2Thermal & 0.291 & 0.276 & 0.224 & 0.370 & 0.290 \\
        MassMIND & 0.454 & 0.310 & 0.445 & 0.488 & 0.424 \\
    \hlineB{3}
    \end{tabular}
    }
\end{table}

\subsubsection{Network Pretraining Ablation}
We evaluate the RGB-pretrained networks (Sec.~\ref{sec:method_pretrain}) on our thermal data in absence of online SSL (Table \ref{table:pretrain-ablation}).  PCA channel mixing outperforms others, possibly because it can modulate the amount of image detail shown However, we leave a thorough investigation and explanation of this observation for future work. RGB-T image translation does not perform well likely because it had limited access to target domain data and introduced numerous structural artifacts that affected segmentation training. 

\subsection{UAV Embedded System Benchmarks}
To demonstrate deploying our algorithm in real-time on UAV hardware, we implement our algorithm in ROS and test with bagfiles on an Nvidia Jetson AGX Orin. 

\subsubsection{ROS architecture} 
\begin{figure}
    \centering
    \includegraphics[width=\columnwidth]{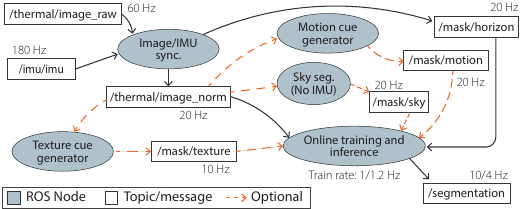}
    \caption{ROS architecture for real-time online learning and water segmentation on a Nvidia Jetson AGX Orin.}
    \label{fig:rqt}
\end{figure}

A node (Fig.~\ref{fig:rqt}) processes incoming thermal imagery (Sec.~\ref{sec:method_preprocess}) and estimates horizon line based on IMU readings or segments the sky using Fast-SCNN if IMU is unavailable. Texture- and/or motion-based labeling nodes generate segmentation labels in parallel. Thermal images and labels are cached in a buffer and training begins once the buffer is full. A third inference network segments incoming images continuously and receives weights from the training network after each online SSL cycle. 
\subsubsection{Computation Benchmarks}
We benchmarked our system using training parameters from Sec.~\ref{section:online-training-params} and list component frequencies in Fig.~\ref{fig:rqt}. Texture- and motion-based adaptation perform online updates at 1 and 1.2~Hz respectively. Actual training takes 0.5~s for 8 iterations with the rest of the time spent filling the training buffer. The inference network produces segmentations at 10 and 4~Hz when using texture and motion cues respectively. Code optimization, better parallelization strategies, and lower online SSL update rates should allow us to attain closer to 15-20~Hz. Overall, we find these metrics to be suitable to enable our future work in nighttime navigation and planning in near-shore areas, as well as other work in bathymetry.

\section{CONCLUSION}

We presented a CNN-based thermal water segmentation algorithm that provides UAVs operating in near-shore environments with nighttime capabilities. We demonstrated that our online SSL approach with simple water cues can achieve strong and consistent results in the aerial setting despite the lack of aerial thermal data. Furthermore, we showed that our method is superior and more robust compared to fully-supervised networks trained on existing thermal data. This work can enable thermal vision-based UAV science missions in near-shore settings for tasks such as bathymetry and coastline mapping. In the future, we look to use this to assist UAV navigation and planning in near-shore environments, and to help curate a larger, aerial thermal near-shore dataset to enable fully-supervised training.

\section*{ACKNOWLEDGMENT}
We thank K. Koetje, K. Brodie, T. Hesser, T. Almeida, T. Cook, and R. Beach for their helpful discussions and thank N. Spore with the U.S. Army ERDC-CHL Field Research Facility for collecting the data sequences at Duck, NC.

\bibliography{refs}

\begin{thebibliography}{10}
\providecommand{\url}[1]{#1}
\csname url@rmstyle\endcsname
\providecommand{\newblock}{\relax}
\providecommand{\bibinfo}[2]{#2}
\providecommand\BIBentrySTDinterwordspacing{\spaceskip=0pt\relax}
\providecommand\BIBentryALTinterwordstretchfactor{4}
\providecommand\BIBentryALTinterwordspacing{\spaceskip=\fontdimen2\font plus
\BIBentryALTinterwordstretchfactor\fontdimen3\font minus
  \fontdimen4\font\relax}
\providecommand\BIBforeignlanguage[2]{{%
\expandafter\ifx\csname l@#1\endcsname\relax
\typeout{** WARNING: IEEEtran.bst: No hyphenation pattern has been}%
\typeout{** loaded for the language `#1'. Using the pattern for}%
\typeout{** the default language instead.}%
\else
\language=\csname l@#1\endcsname
\fi
#2}}

\bibitem{yang2017vision}
J.~Yang, A.~Dani, S.-J. Chung, and S.~Hutchinson, ``Vision-based localization
  and robot-centric mapping in riverine environments,'' \emph{J. of Field
  Robotics}, vol.~34, no.~3, pp. 429--450, 2017.

\bibitem{8726410}
K.~L. Brodie, B.~L. Bruder, R.~K. Slocum, and N.~J. Spore, ``Simultaneous
  mapping of coastal topography and bathymetry from a lightweight multicamera
  uas,'' \emph{{IEEE} Trans. Geosci. Remote Sensing}, vol.~57, no.~9, pp.
  6844--6864, 2019.

\bibitem{bb_iros_2019}
B.~Bovcon, J.~Muhovi{\v{c}}, J.~Per{\v{s}}, and M.~Kristan, ``The mastr1325
  dataset for training deep usv obstacle detection models,'' in \emph{Proc.
  {IEEE}/{RSJ} Int. Conf. Intell. Robots and Syst.}\hskip 1em plus 0.5em minus
  0.4em\relax IEEE, 2019.

\bibitem{yao2021shorelinenet}
L.~Yao, D.~Kanoulas, Z.~Ji, and Y.~Liu, ``Shorelinenet: an efficient deep
  learning approach for shoreline semantic segmentation for unmanned surface
  vehicles,'' in \emph{Proc. {IEEE}/{RSJ} Int. Conf. Intell. Robots and
  Syst.}\hskip 1em plus 0.5em minus 0.4em\relax IEEE, 2021, pp. 5403--5409.

\bibitem{bovcon2018stereo}
B.~Bovcon, J.~Per{\v{s}}, M.~Kristan, \emph{et~al.}, ``Stereo obstacle
  detection for unmanned surface vehicles by imu-assisted semantic
  segmentation,'' \emph{Robot. and Auton. Syst.}, vol. 104, pp. 1--13, 2018.

\bibitem{msuda}
Y.-H. Kim, U.~Shin, J.~Park, and I.~S. Kweon, ``Ms-uda: Multi-spectral
  unsupervised domain adaptation for thermal image semantic segmentation,''
  \emph{IEEE Robotics and Automation Letters}, vol.~6, no.~4, pp. 6497--6504,
  2021.

\bibitem{akkaya2021self}
I.~B. Akkaya, F.~Altinel, and U.~Halici, ``Self-training guided adversarial
  domain adaptation for thermal imagery,'' in \emph{Proc. {IEEE} Conf. Comput.
  Vis. Pattern Recog.}, 2021, pp. 4322--4331.

\bibitem{li2020segmenting}
C.~Li, W.~Xia, Y.~Yan, B.~Luo, and J.~Tang, ``Segmenting objects in day and
  night: Edge-conditioned cnn for thermal image semantic segmentation,''
  \emph{IEEE Transactions on Neural Networks and Learning Systems}, vol.~32,
  no.~7, pp. 3069--3082, 2020.

\bibitem{chen2022light}
J.~Chen, Z.~Liu, D.~Jin, Y.~Wang, F.~Yang, and X.~Bai, ``Light transport
  induced domain adaptation for semantic segmentation in thermal infrared urban
  scenes,'' \emph{{IEEE} Trans. Intell. Transport. Syst.}, vol.~23, no.~12, pp.
  23\,194--23\,211, 2022.

\bibitem{qian2020sparse}
X.~Qian, M.~Zhang, and F.~Zhang, ``Sparse gans for thermal infrared image
  generation from optical image,'' \emph{IEEE Access}, vol.~8, pp.
  180\,124--180\,132, 2020.

\bibitem{zhang2018synthetic}
L.~Zhang, A.~Gonzalez-Garcia, J.~Van De~Weijer, M.~Danelljan, and F.~S. Khan,
  ``Synthetic data generation for end-to-end thermal infrared tracking,''
  \emph{{IEEE} Trans. Image Processing}, vol.~28, no.~4, pp. 1837--1850, 2018.

\bibitem{ros}
M.~Quigley, B.~Gerkey, K.~Conley, J.~Faust, T.~Foote, J.~Leibs, \emph{et~al.},
  ``Ros: an open-source robot operating system,'' in \emph{{ICRA} Workshop on
  Open Source Software}, Kobe, Japan, May 2009.

\bibitem{rankin2006daytime}
A.~L. Rankin, L.~H. Matthies, and A.~Huertas, ``Daytime water detection by
  fusing multiple cues for autonomous off-road navigation,'' in
  \emph{Transformational Science And Technology For The Current And Future
  Force}.\hskip 1em plus 0.5em minus 0.4em\relax World Scientific, 2006, pp.
  177--184.

\bibitem{RankinWater}
A.~Rankin and L.~Matthies, ``Daytime water detection based on color
  variation,'' in \emph{Proc. {IEEE}/{RSJ} Int. Conf. Intell. Robots and
  Syst.}, 2010, pp. 215--221.

\bibitem{meier2021river}
K.~Meier, S.-J. Chung, and S.~Hutchinson, ``River segmentation for autonomous
  surface vehicle localization and river boundary mapping,'' \emph{J. of Field
  Robotics}, vol.~38, no.~2, pp. 192--211, 2021.

\bibitem{achar2011self}
S.~Achar, B.~Sankaran, S.~Nuske, S.~Scherer, and S.~Singh, ``Self-supervised
  segmentation of river scenes,'' in \emph{Proc. {IEEE} Int. Conf. Robot. and
  Automation}.\hskip 1em plus 0.5em minus 0.4em\relax IEEE, 2011, pp.
  6227--6232.

\bibitem{kristan2015fast}
M.~Kristan, V.~S. Kenk, S.~Kova{\v{c}}i{\v{c}}, and J.~Per{\v{s}}, ``Fast
  image-based obstacle detection from unmanned surface vehicles,'' \emph{IEEE
  Trans. on Cybernetics}, vol.~46, no.~3, pp. 641--654, 2015.

\bibitem{9197194}
B.~Bovcon and M.~Kristan, ``A water-obstacle separation and refinement network
  for unmanned surface vehicles,'' in \emph{Proc. {IEEE} Int. Conf. Robot. and
  Automation}, 2020, pp. 9470--9476.

\bibitem{nirgudkar2021beyond}
S.~Nirgudkar and P.~Robinette, ``Beyond visible light: Usage of long wave
  infrared for object detection in maritime environment,'' in \emph{Int. Conf.
  on Advanced Robotics (ICAR)}.\hskip 1em plus 0.5em minus 0.4em\relax IEEE,
  2021, pp. 1093--1100.

\bibitem{nirgudkar2023massmind}
S.~Nirgudkar, M.~DeFilippo, M.~Sacarny, M.~Benjamin, and P.~Robinette,
  ``Massmind: Massachusetts maritime infrared dataset,'' \emph{Int. J. Robot.
  Res.}, vol.~42, no. 1-2, pp. 21--32, 2023.

\bibitem{isola2017image}
P.~Isola, J.-Y. Zhu, T.~Zhou, and A.~A. Efros, ``Image-to-image translation
  with conditional adversarial networks,'' in \emph{Proc. {IEEE} Conf. Comput.
  Vis. Pattern Recog.}, 2017.

\bibitem{gan2023unsupervised}
L.~Gan, C.~Lee, and S.-J. Chung, ``Unsupervised rgb-to-thermal domain
  adaptation via multi-domain attention network,'' in \emph{Proc. {IEEE} Int.
  Conf. Robot. and Automation}.\hskip 1em plus 0.5em minus 0.4em\relax IEEE,
  2023, pp. 6014--6020.

\bibitem{5650300}
S.~Zhou and K.~Iagnemma, ``Self-supervised learning method for unstructured
  road detection using fuzzy support vector machines,'' in \emph{Proc.
  {IEEE}/{RSJ} Int. Conf. Intell. Robots and Syst.}, 2010, pp. 1183--1189.

\bibitem{dahlkamp2006self}
H.~Dahlkamp, A.~Kaehler, D.~Stavens, S.~Thrun, and G.~R. Bradski,
  ``Self-supervised monocular road detection in desert terrain.'' in
  \emph{Proc. Robot.: Sci. Syst. Conf.}, vol.~38.\hskip 1em plus 0.5em minus
  0.4em\relax Philadelphia, 2006.

\bibitem{8486472}
J.~Cho, Y.~Kim, H.~Jung, C.~Oh, J.~Youn, and K.~Sohn, ``Multi-task
  self-supervised visual representation learning for monocular road
  segmentation,'' in \emph{2018 IEEE International Conference on Multimedia and
  Expo (ICME)}, 2018, pp. 1--6.

\bibitem{schmid2022self}
R.~Schmid, D.~Atha, F.~Sch{\"o}ller, S.~Dey, S.~Fakoorian, K.~Otsu, B.~Ridge,
  M.~Bjelonic, L.~Wellhausen, M.~Hutter, \emph{et~al.}, ``Self-supervised
  traversability prediction by learning to reconstruct safe terrain,'' in
  \emph{Proc. {IEEE}/{RSJ} Int. Conf. Intell. Robots and Syst.}\hskip 1em plus
  0.5em minus 0.4em\relax IEEE, 2022, pp. 12\,419--12\,425.

\bibitem{chen2020simple}
T.~Chen, S.~Kornblith, M.~Norouzi, and G.~Hinton, ``A simple framework for
  contrastive learning of visual representations,'' in \emph{International
  conference on machine learning}.\hskip 1em plus 0.5em minus 0.4em\relax PMLR,
  2020, pp. 1597--1607.

\bibitem{chen2021exploring}
X.~Chen and K.~He, ``Exploring simple siamese representation learning,'' in
  \emph{Proc. {IEEE} Conf. Comput. Vis. Pattern Recog.}, 2021, pp.
  15\,750--15\,758.

\bibitem{araslanov2021self}
N.~Araslanov and S.~Roth, ``Self-supervised augmentation consistency for
  adapting semantic segmentation,'' in \emph{Proc. {IEEE} Conf. Comput. Vis.
  Pattern Recog.}, 2021, pp. 15\,384--15\,394.

\bibitem{zhan2019autonomous}
W.~Zhan, C.~Xiao, Y.~Wen, C.~Zhou, H.~Yuan, S.~Xiu, Y.~Zhang, X.~Zou, X.~Liu,
  and Q.~Li, ``Autonomous visual perception for unmanned surface vehicle
  navigation in an unknown environment,'' \emph{Sensors}, vol.~19, no.~10, p.
  2216, 2019.

\bibitem{daftry2018online}
S.~Daftry, Y.~Agrawal, and L.~Matthies, ``Online self-supervised long-range
  scene segmentation for {MAVs},'' in \emph{Proc. {IEEE}/{RSJ} Int. Conf.
  Intell. Robots and Syst.}\hskip 1em plus 0.5em minus 0.4em\relax IEEE, 2018,
  pp. 5194--5199.

\bibitem{pizer1987adaptive}
S.~M. Pizer, E.~P. Amburn, J.~D. Austin, R.~Cromartie, A.~Geselowitz, T.~Greer,
  \emph{et~al.}, ``Adaptive histogram equalization and its variations,''
  \emph{Computer vision, graphics, and image processing}, vol.~39, no.~3, pp.
  355--368, 1987.

\bibitem{lin2017feature}
T.-Y. Lin, P.~Doll{\'a}r, R.~Girshick, K.~He, B.~Hariharan, and S.~Belongie,
  ``Feature pyramid networks for object detection,'' in \emph{Proc. {IEEE}
  Conf. Comput. Vis. Pattern Recog.}, 2017, pp. 2117--2125.

\bibitem{howard2019searching}
A.~Howard, M.~Sandler, G.~Chu, L.-C. Chen, B.~Chen, M.~Tan, W.~Wang, Y.~Zhu,
  R.~Pang, \emph{et~al.}, ``Searching for mobilenetv3,'' in \emph{Proc. {IEEE}
  Int. Conf. Comput. Vis.}, 2019, pp. 1314--1324.

\bibitem{zhou2018semantic}
B.~Zhou, H.~Zhao, X.~Puig, T.~Xiao, S.~Fidler, A.~Barriuso, and A.~Torralba,
  ``Semantic understanding of scenes through the ade20k dataset,'' \emph{Int.
  J. Comput. Vis.}, 2018.

\bibitem{caesar2018coco}
H.~Caesar, J.~Uijlings, and V.~Ferrari, ``Coco-stuff: Thing and stuff classes
  in context,'' in \emph{Proc. {IEEE} Conf. Comput. Vis. Pattern Recog.}, 2018,
  pp. 1209--1218.

\bibitem{8258373}
L.~Lopez-Fuentes, C.~Rossi, and H.~Skinnemoen, ``River segmentation for flood
  monitoring,'' in \emph{2017 IEEE Int. Conf. on Big Data}, 2017, pp.
  3746--3749.

\bibitem{park2020cut}
T.~Park, A.~A. Efros, R.~Zhang, and J.-Y. Zhu, ``Contrastive learning for
  unpaired image-to-image translation,'' in \emph{Proc. European Conf. Comput.
  Vis.}, 2020.

\bibitem{achanta2010slic}
R.~Achanta, A.~Shaji, K.~Smith, A.~Lucchi, P.~Fua, and S.~S{\"u}sstrunk, ``Slic
  superpixels,'' Tech. Rep., 2010.

\bibitem{lowe1999object}
D.~G. Lowe, ``Object recognition from local scale-invariant features,'' in
  \emph{Proc. {IEEE} Int. Conf. Comput. Vis.}, vol.~2, 1999, pp. 1150--1157.

\bibitem{orb}
E.~Rublee, V.~Rabaud, K.~Konolige, and G.~Bradski, ``Orb: An efficient
  alternative to sift or surf,'' in \emph{Proc. {IEEE} Int. Conf. Comput.
  Vis.}, 2011, pp. 2564--2571.

\bibitem{ransac}
\BIBentryALTinterwordspacing
M.~A. Fischler and R.~C. Bolles, ``Random sample consensus: A paradigm for
  model fitting with applications to image analysis and automated
  cartography,'' \emph{Commun. ACM}, vol.~24, no.~6, p. 381–395, jun 1981.
  [Online]. Available: \url{https://doi.org/10.1145/358669.358692}
\BIBentrySTDinterwordspacing

\bibitem{farneback2003two}
G.~Farneb{\"a}ck, ``Two-frame motion estimation based on polynomial
  expansion,'' in \emph{Image Analysis: 13th Scandinavian Conf., SCIA 2003
  Halmstad, Sweden, June 29--July 2, 2003 Proceedings 13}.\hskip 1em plus 0.5em
  minus 0.4em\relax Springer, 2003, pp. 363--370.

\bibitem{poudel2019fast}
R.~P. Poudel, S.~Liwicki, and R.~Cipolla, ``Fast-scnn: Fast semantic
  segmentation network,'' \emph{arXiv preprint arXiv:1902.04502}, 2019.

\bibitem{hwang2015multispectral}
S.~Hwang, J.~Park, N.~Kim, Y.~Choi, and I.~S. Kweon, ``Multispectral pedestrian
  detection: Benchmark dataset and baselines,'' in \emph{Proc. {IEEE} Conf.
  Comput. Vis. Pattern Recog.}, 2015.

\bibitem{ZHANG_2020_ICIP}
H.~Zhang, E.~Fromont, S.~Lef{\`e}vre, and B.~Avignon, ``Multispectral fusion
  for object detection with cyclic fuse-and-refine blocks,'' in \emph{Proc.
  Int. Conf. Image Process.}, Abu Dhabi, United Arab Emirates, Oct. 2020, pp.
  1--5.

\end{thebibliography}

\end{document}